\documentclass{article}

\PassOptionsToPackage{numbers, compress}{natbib}  

 \usepackage[dblblindworkshop, final]{neurips_2025}
\workshoptitle{The Second Workshop on GenAI for Health: Potential, Trust, and Policy Compliance}

\usepackage[utf8]{inputenc}  
\usepackage[T1]{fontenc}     

\usepackage{microtype}       
\usepackage{xcolor}          
\usepackage{enumitem}        
\usepackage{tcolorbox}       

\usepackage{amsmath}         
\usepackage{amsfonts}        
\usepackage{nicefrac}        
\usepackage{amsthm}          
\usepackage{siunitx}         

\usepackage{booktabs}        
\usepackage{makecell}        
\usepackage{multirow}        
\usepackage{tabularx}        
\usepackage{array}           
\usepackage{ragged2e}

\newcolumntype{C}[1]{>{\centering\arraybackslash}p{#1}}

\usepackage{graphicx}        

\usepackage{hyperref}        
\usepackage{url}             

\usepackage{listings}        
\usepackage[multiple]{footmisc}

\title{Customizing Open‑Source LLMs for Quantitative Medication Attribute Extraction across Heterogeneous EHR Systems}

\author{
  Zhe Fei\thanks{These authors contributed equally.} \textsuperscript{ ,}\thanks{Department of Statistics, UC Riverside, Riverside, CA, 92521}
  \And
  Mehmet Yigit Turali\footnotemark[1] \textsuperscript{ ,}\thanks{Department of Electrical and Computer Engineering, UCLA, Los Angeles, CA, 90095}
  \And
  Shreyas Rajesh\footnotemark[1] \textsuperscript{ ,}\footnotemark[3]
  \AND 
  Xinyang Dai\thanks{Department of Psychiatry and Biobehavioral Sciences, UCLA, Los Angeles, CA 90024}
  \And
  Huyen Pham\footnotemark[4]
  \And
Pavan Holur\footnotemark[3]
  \And
  Yuhui Zhu\footnotemark[4]
  \AND 
  Larissa Mooney\footnotemark[4]
  \And
  Yih-Ing Hser\footnotemark[4]
  \And
  Vwani Roychowdhury\footnotemark[3] 
}

\begin{document}

\makeatletter
\providecommand{\@trackname}{}
\makeatother
\maketitle

\begin{abstract}
 Harmonizing medication data across Electronic Health Record (EHR) systems is a persistent barrier to monitoring medications for opioid use disorder (MOUD). In heterogeneous EHR systems, key prescription attributes are scattered across differently formatted fields and free‐text notes. We present a practical framework that customizes open-source large language models (LLMs), including Llama, Qwen, Gemma, and MedGemma, to extract a unified set of MOUD prescription attributes (prescription date, drug name, duration, total quantity, daily quantity, and refills) from heterogeneous, site-specific data and compute a standardized metric of medication coverage, \emph{MOUD days}, per patient. Our pipeline processes records directly in a fixed JSON schema, followed by lightweight normalization and cross-field consistency checks. We evaluate the system on prescription-level EHR data from five clinics in a national OUD study (25{,}605 records from 1{,}257 patients), using a previously annotated benchmark of 10{,}369 records (776 patients) as the ground truth. Performance is reported as coverage (share of records with a valid, matchable output) and record-level exact-match accuracy. Larger models perform best overall: Qwen2.5-32B achieves \textbf{93.4\%} coverage with \textbf{93.0\%} exact-match accuracy across clinics, and MedGemma-27B attains \textbf{93.1\%}/\textbf{92.2\%}. A brief error review highlights three common issues and fixes: imputing missing dosage fields using within-drug norms, handling monthly/weekly injectables (e.g., Vivitrol) by setting duration from the documented schedule, and adding unit checks to prevent mass units (e.g., ``250 g'') from being misread as daily counts. By removing brittle, site-specific ETL and supporting local, privacy-preserving deployment, this approach enables consistent cross-site analyses of MOUD exposure, adherence, and retention in real-world settings.
\end{abstract}

\section{Introduction}
\label{sec:intro}

The opioid crisis remains a major public health issue impacting communities across the United States, with over 105,000 overdose deaths recorded between December 2022 and January 2023 \citep{Ahmad2023Provisional}. While the opioid crisis has not been limited to a specific region, rural communities have been particularly hit hard \citep{Mack2017Illicit}. Medications for opioid use disorder (MOUD) have been identified as an effective treatment approach in reducing opioid use \citep{Connery2015MATReview,Volkow2014NEJM}. However, they tend to be significantly underutilized, especially in rural communities \citep{Amiri2021ValueHealth,Amiri2021DAD,Showers2021JRuralMH}. Individuals with OUD in rural communities face significant challenges in accessing MOUD treatment, largely due to geographical isolation, limited transportation infrastructure, and a shortage of providers \citep{Lister2020AJDAA,Bommersbach2023IJDP}.

{Effective OUD treatment monitoring and quality improvement require systematic analysis of medication patterns   across healthcare systems. However, this analysis is severely hampered by heterogeneity across Electronic Health Record (EHR) systems, critical MOUD prescription data are scattered across differently formatted structured fields and unstructured clinical notes, making cross-clinic comparisons extremely tedious. Traditional ETL (Extract Transform Load) approaches require custom mappings for each EHR system and are highly brittle: even minor changes in field names, data formats, or clinical documentation patterns can break extraction pipelines, creating substantial maintenance burdens and limiting scalability. 
   The emergence of large language models (LLMs)\cite{openai_gpt-4_2024,team2024gemini,qwen3} offers a promising solution to this challenge, driven by their impressive instruction-following and natural language understanding capabilities. Recent advances have also enhanced their ability to reliably generate structured outputs, allowing them to return data that conforms to a predefined format such as JSON. Utilizing these dual strengths, we developed a framework that leverages open-source LLMs (Qwen, Llama) to extract and standardize MOUD prescription attributes—prescription date, drug name, duration, quantities, and refills—from five clinics' disparate EHR systems. The system computes standardized "MOUD days" (medication coverage duration) for each prescription, enabling consistent cross-site analyses of treatment patterns, retention, and adherence without site-specific ETL development.}

{We demonstrate our system's capabilities using EHR data from five rural clinics participating in a national OUD study (CTN-0102C supported by National Drug Abuse Treatment Clinical Trials Network). Section \ref{sec:methods} details our system architecture and implementation, as well as the data used to evaluate our framework. Section \ref{sec:results} presents extraction performance across different open-source models and clinic sites. Section \ref{sec:disc} discusses deployment considerations and limitations in future real-world clinical settings.}



\section{Methods}
\label{sec:methods}



\begin{figure}[htbp]
    \centering
    \includegraphics[width=\textwidth]{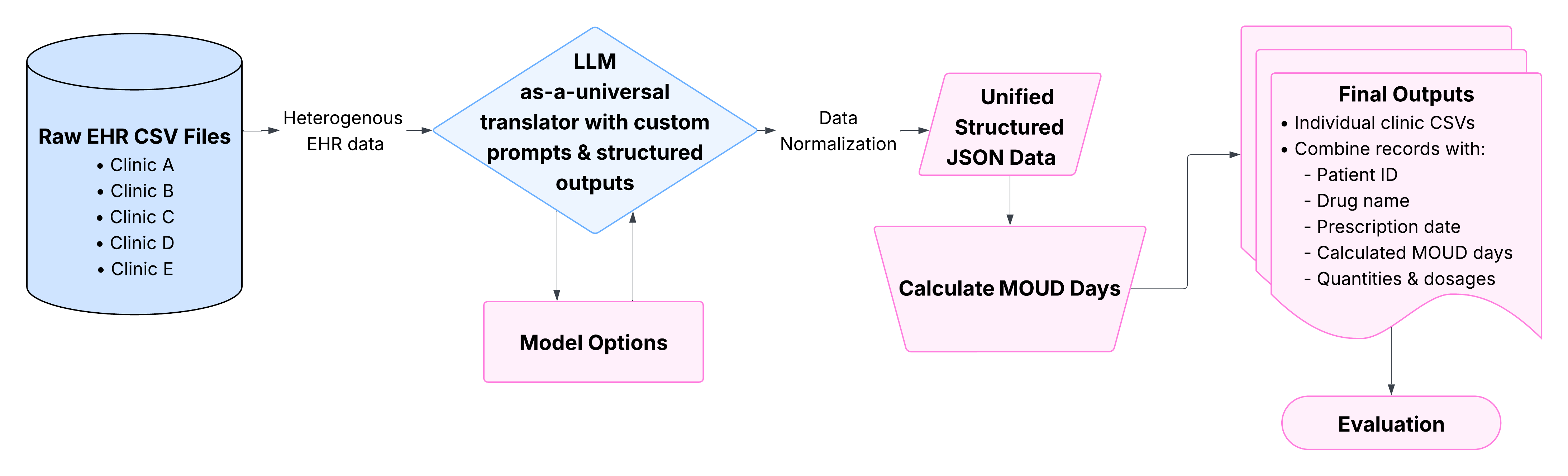}
    \caption{An overview of the framework for MOUD extraction and calculation. The system processes heterogeneous raw EHR data, uses an LLM to translate it into a unified JSON structure, calculates MOUD days, and evaluates the final output against a ground truth dataset.}
    \label{fig:pipeline}
\end{figure}

Our system is designed to extract and harmonize MOUD prescription information from heterogeneous EHR data sources. \textbf{Figure \ref{fig:pipeline} provides a high-level overview of this framework.} The process begins by ingesting raw, heterogeneous EHR data from various clinics. At its core, an instruction-tuned LLM acts as a universal translator, converting this diverse data into a unified, structured JSON format. This standardized output then undergoes post-processing to calculate the final MOUD days metric. Finally, the results are evaluated against a manually annotated ground truth dataset to measure performance in terms of accuracy and coverage. The framework consists of three main stages, detailed below: (1) data preparation, (2) LLM-based unified output extraction, and (3) MOUD computation and evaluation.

\subsection{Data Preparation}
\label{subsec:data}



To evaluate our framework, we used prescription-level EHR data from five clinics participating in a multi-site national study on OUD, denoted as A -- E. While the full dataset comprises \textbf{25,605 records} from \textbf{1,257 unique patients}, a benchmark ground truth set from the feasibility study period was established. This subset, consisting of \textbf{10,369 records} from \textbf{776 unique patients}, was manually annotated by UCLA medical professionals to extract the canonical values for MOUD days calculation. This annotated set serves as the gold standard for our quantitative evaluation, and its distribution is described in Table \ref{tab:moud_counts_examples}.

Across all sites, we targeted a common set of medication attributes to enable standardized calculation of "MOUD days": \textbf{clinic name, patient ID, prescription date, drug name, duration, total prescribed quantity, daily quantity, and number of refills}. These essential attributes were found in heterogeneous structured columns and, for some clinics, unstructured notes (Table \ref{tab:moud_counts_examples}).

\begin{table}[htbp]
\centering
\small
\caption{Ground truth data and example raw EHR fields by clinic (A--E).}
\label{tab:moud_counts_examples}
\renewcommand{\arraystretch}{1.15}
\begin{tabularx}{\textwidth}{C{0.9cm} C{1.2cm} C{1cm} >{\raggedright\arraybackslash}X}
\toprule
\textbf{Clinic} & \textbf{Annotated Records} & \textbf{Patients} & \textbf{Example raw EHR entries (name : value)} \\
\midrule

A & 7 & 7 &
\texttt{BRAND\_NAME}:\ \texttt{BUPRENORPHINE-NALOXONE};\qquad
\texttt{GENERIC\_NAME}:\ \texttt{buprenorphine HCl/naloxone HCl};\qquad
\texttt{PRESCRIPITION\_DATE}:\ \texttt{9/15/21};\qquad
\texttt{ROUTE\_OF\_ADMINISTRATION}:\ \texttt{SUBLINGUAL};\qquad
\texttt{UNIT\_DOSE}:\ \texttt{2 mg-0.5 mg};\qquad
\texttt{PRESCRIBED\_QUANTITY}:\ \texttt{1};\qquad
\texttt{DOSAGE\_INSTRUCTIONS}:\ \texttt{place 2 tablet by sublingual route every day \dots\ (2 tabs in am and 1 in pm)};\qquad
\texttt{MEDICATION\_INDICATION1}:\ \texttt{F11.11};\qquad
\texttt{ORIGINAL\_REFILLS}:\ \texttt{0};\qquad
\texttt{DATE\_STOPPED}:\ \texttt{1/27/22};\qquad
\texttt{RECORDID}:\ \texttt{123835}
\\[0.25em]

B & 5295 & 410 &
\texttt{record\_num}:\ \texttt{322853};\qquad
\texttt{epic\_medication\_id}:\ \texttt{120111686};\qquad
\texttt{epic\_medication\_name}:\ \texttt{BUPRENORPHINE HCL-NALOXONE HCL 8-2 MG SL SUBL};\qquad
\texttt{med\_route}:\ \texttt{Sublingual};\qquad
\texttt{dose\_unit}:\ \texttt{tablet};\qquad
\texttt{dose\_instructions}:\ \texttt{Place 1 tablet under the tongue every 8 hours as needed for up to 28 days.};\qquad
\texttt{frequency}:\ \texttt{EVERY 8 HOURS PRN};\qquad
\texttt{quantity}:\ \texttt{84 tablet};\qquad
\texttt{refill}:\ \texttt{0};\qquad
\texttt{prescription\_date}:\ \texttt{10/29/19}
\\[0.25em]

C & 4402 & 286 &
\texttt{RecordID}:\ \texttt{102724};\qquad
\texttt{DrugDescription}:\ \texttt{Suboxone 2-0.5 mg film};\qquad
\texttt{PrescribedDate}:\ \texttt{6/17/21};\qquad
\texttt{SUMMARY}:\ \texttt{Suboxone 2-0.5 mg film};\qquad
\texttt{ROUTE}:\ \texttt{UNDER TONGUE};\qquad
\texttt{INSTRUCTIONS}:\ \texttt{Place 0.25 strip under tongue once a day For chronic OUD, XS2110928};\qquad \texttt{Refills}:\ \texttt{0};\qquad
\texttt{DOSE\_UNIT}:\ \texttt{strip};\qquad
\texttt{PrescribedQuantity}:\ \texttt{7 film};\qquad
\texttt{DoseQuantity}:\ \texttt{NULL}
\\[0.25em]

D & 139 & 18 &
\texttt{RecordNumber}:\ \texttt{926195};\qquad
\texttt{Code}:\ \texttt{657570300};\qquad
\texttt{Description}:\ \texttt{Vivitrol 380 mg suspension, extended rel recon};\qquad
\texttt{PrescriptionDate}:\ \texttt{8/7/23};\qquad
\texttt{UnitDosage}:\ \texttt{ };\\[-0.5em]
& & & 
\texttt{DosageInstructions}:\ \texttt{INJECT 380 MG INTRAMUSCULARLY EVERY FOUR WEEKS \dots};\qquad
\texttt{DoseQuantity}:\ \texttt{1 each};\qquad
\texttt{NumberOfRefillsAuthorized}:\ \texttt{2}
\\[0.25em]

E & 426 & 27 &
\texttt{RecordID}:\ \texttt{114147};\qquad
\texttt{PrescribedDate}:\ \texttt{6/22/21};\qquad
\texttt{DrugDescription}:\ \texttt{Suboxone 8-2 mg film};\qquad
\texttt{SUMMARY}:\ \texttt{pt missed appt today};\qquad
\texttt{ROUTE}:\ \texttt{UNDER TONGUE};\qquad
\texttt{Refills}:\ \texttt{0};\qquad
\texttt{INSTRUCTIONS}:\ \texttt{Dissolve 1 film under tongue once a day};\qquad
\texttt{DOSE\_UNIT}:\ \texttt{film};\qquad
\texttt{PrescribedQuantity}:\ \texttt{28 film}
\\

\midrule
\textbf{Total} & \textbf{10369} & \textbf{776} & \\
\bottomrule
\end{tabularx}
\end{table}

\subsection{LLM-based Unified Output Extraction}
\label{subsec:system}

Our extraction framework leverages instruction-tuned LLMs as a universal translator, converting heterogeneous EHR data into a common, unified schema. The core of our approach utilizes constrained generation: instead of parsing unstructured text output, we force the model's generation to directly conform to a predefined Pydantic-based JSON schema \cite{Colvin_Pydantic_Validation_2025}. This method guarantees syntactically correct, structured data and has been shown to improve task performance by eliminating parsing errors \cite{outlines}. The structured output pipeline enforces strict adherence to schemas containing required fields (patient id, prescription date, drug name, etc.) crucial for computing MOUD days. We detail the prompts used across each clinic and the unified output schema in Appendix \ref{app:sec:prescription_prompts}.

To implement this strategy, we evaluated a comprehensive suite of open-source LLMs. Due to privacy regulations and HIPAA compliance requirements, only models that could be deployed locally were considered, excluding closed-source APIs such as GPT-4 \cite{openai_gpt-4_2024} or Gemini \cite{team2024gemini}. The selected models, chosen to assess performance-efficiency trade-offs, included Qwen2.5 (32B) \cite{yang2024qwen2}, Qwen3 (4B, 8B, 32B)\cite{yang_qwen3_2025}, Gemma (4B, 7B, 27B), and MedGemma (4B, 27B). Initial experiments with sub-1B models revealed significant difficulties in generating reliable structured outputs, leading us to focus on these larger models.

For efficient and scalable inference, all models were deployed using the vLLM framework \cite{vllm} on a single node with 4 NVIDIA A6000 GPUs (48GB VRAM each). We employed model quantization strategies \cite{jacob2017quantizationtrainingneuralnetworks, gptq} to manage memory usage while maintaining inference quality. Tensor parallel processing was used across both GPUs to optimize throughput, with GPU memory utilization capped at 80\% for stability. Detailed model parameters and inference settings are provided in Appendix \ref{app:sec:model_specs} and \ref{app:sec:inference_config}.

\subsection{Post-Processing and MOUD Days Calculation}
\label{subsec:moud}
The JSON output from the LLM is parsed and subjected to a series of post-processing steps to ensure data quality and consistency. These include:
\begin{itemize}
    \item \textbf{Type Normalization:} Casting extracted values to their correct data types (e.g., dates, integers) for ease of comparison with the ground truth.
    \item \textbf{Rule-Based Validation:} Applying rules to flag logical inconsistencies and remove duplicates. For example, a cross-field check ensures that \textit{total quantity} is not less than \textit{daily quantity}, and additional checks are implemented to filter out nonsensical and null values.
    \item \textbf{MOUD Days Calculation:} The primary outcome, "MOUD days", representing the total potential medication coverage from a prescription and its refills, is computed. A hierarchical logic is used which prioritizes the explicitly extracted duration. If duration is not available, it is derived from the quantity and dosage information. The total MOUD days are calculated for each record according to the following formula:
\end{itemize}

$$
\text{MOUD\_days} = (\text{number\_of\_refills} + 1) \times
\begin{cases}
  \text{duration},  \text{if duration is provided;} \\
  \frac{\text{total\_quantity}}{\text{daily\_quantity}},  \text{if duration is missing.}
\end{cases}
$$

\subsection{Evaluation Pipeline}
\label{subsec:evalpipeline}
We evaluated our framework's performance and flexibility by integrating several leading open-source LLMs as its core information extraction engine. The output from each model was evaluated against the manually annotated ground truth data. To assess robustness against real-world data variations, the evaluation was conducted on a per-clinic basis, reflecting the distinct EHR data structures at each site. Performance was measured using the following quantitative metrics:
\begin{itemize}
    \item \textbf{Coverage (\%):} The percentage of ground truth records for which the model successfully generated a parsable output that could be matched using a composite key of \textbf{clinic name, patient ID, prescription date, and drug name}. This metric measures the system's ability to successfully process each input record before its accuracy is assessed.
    \item \textbf{Record-Level Exact Match Accuracy (\%):} For the records successfully covered, this is the percentage where all five extracted attributes (as detailed in Subsection \ref{subsec:data}) and MOUD days perfectly match the ground truth values after normalization.
\end{itemize}

{In addition to these metrics, we performed a qualitative error analysis to identify common failure modes, such as models struggling with ambiguous phrasing in clinical notes or complex dosage instructions. We further present these in Section \ref{sec:results}}

\begin{table}[htbp]
\centering
\small 
\setlength{\tabcolsep}{2.8pt} 
\caption{\textbf{Performance of LLMs on MOUD Extraction}. Coverage (Cov.) and Exact Match Accuracy (Acc.) are in percentages (\%). Clinic names are abbreviated. Full model details are in Appendix \ref{app:sec:model_specs}.}
\label{tab:main_results_fit}
\begin{tabular}{@{}l cc cc cc cc cc cc@{}}
\toprule
& \multicolumn{2}{c}{A} & \multicolumn{2}{c}{B} & \multicolumn{2}{c}{C} & \multicolumn{2}{c}{D} & \multicolumn{2}{c}{E} & \multicolumn{2}{c}{\textbf{Overall}} \\
\cmidrule(lr){2-3} \cmidrule(lr){4-5} \cmidrule(lr){6-7} \cmidrule(lr){8-9} \cmidrule(lr){10-11} \cmidrule(lr){12-13}
Model & Cov. & Acc. & Cov. & Acc. & Cov. & Acc. & Cov. & Acc. & Cov. & Acc. & Cov. & Acc. \\
\midrule
Qwen2.5 (32B) & 100.00 & 100.00 & 95.98 & 93.51 & 96.32 & 93.04 & 97.12 & 80.74 & 52.35 & 89.69 & 93.4 & 93.0 \\
Qwen3 (32B) & 42.86 & 66.67 & 87.99 & 91.89 & 59.93 & 89.08 & 97.12 & 95.56 & 93.66 & 89.97 & 75.6 & 90.9 \\
Qwen3 (8B) & 100.00 & 100.00 & 89.54 & 92.77 & 51.61 & 69.24 & 27.34 & 94.74 & 42.96 & 95.63 & 69.4 & 85.2 \\
Qwen3 (4B) & 71.43 & 100.00 & 95.43 & 92.12 & 2.20 & 25.77 & 97.12 & 97.78 & 71.60 & 90.16 & 54.0 & 91.0 \\
\midrule
MedGemma (27B) & 85.71 & 50.00 & 93.41 & 93.95 & 96.23 & 90.01 & 96.40 & 98.51 & 76.76 & 92.66 & 93.1 & 92.2 \\
MedGemma (4B) & 85.71 & 83.33 & 48.56 & 81.87 & 14.93 & 18.26 & 96.40 & 97.01 & 44.84 & 32.98 & 34.3 & 68.1 \\
\midrule
Gemma 3 (27B) & 100.00 & 85.71 & 76.81 & 91.76 & 82.01 & 90.08 & 74.82 & 86.54 & 46.01 & 98.98 & 77.0 & 91.1 \\
Gemma 3 (4B) & 100.00 & 85.71 & 81.91 & 83.49 & 50.61 & 68.04 & 93.53 & 65.38 & 19.95 & 81.18 & 65.5 & 78.0 \\
\bottomrule
\end{tabular}
\end{table}

\section{Results}
\label{sec:results}

The performance of the selected open-source LLMs on the MOUD extraction task is presented in Table \ref{tab:main_results_fit}. Our evaluation reveals two primary findings: 1) a significant variation in performance across different clinic datasets, highlighting the challenge of data heterogeneity, and 2) a consistent trend where models achieve high exact-match accuracy on the records they successfully process, but often struggle with overall record coverage.

A stark performance gap was observed between clinics. For \textbf{Clinic B and D}, most models achieved excellent coverage (often >95\%) and high accuracy (>90\%), suggesting that the data formats from these clinics are more amenable to automated extraction. In sharp contrast, a wide variance in performance was seen for \textbf{Clinic A and C}, with some models achieving perfect scores, while others failed completely. For example, \textbf{Qwen2.5 32B} achieved 100\% on both metrics for Clinic A, while \textbf{Qwen3 32B} scored 0\%. \textbf{Clinic E} represented a middle ground, where models like \textbf{Qwen3 32B} achieved high coverage (93.66\%) while others, like \textbf{Gemma3 4B}, were less effective (19.95\%).

When comparing model performance, the larger models generally yielded the best results. \textbf{Qwen2.5 32B} emerged as the most balanced model, achieving the highest overall coverage of \textbf{93.4\%} while maintaining a strong overall accuracy of \textbf{93.0\%}. \textbf{MedGemma (27B)} also performed robustly, with a comparable overall coverage of \textbf{93.1\%} and accuracy of \textbf{92.2\%}. While the smaller \textbf{Qwen3} models (4B and 8B) were competitive on accuracy, they exhibited lower and more inconsistent coverage, with the \textbf{Qwen3 8B} model's overall coverage dropping to \textbf{48.0\%}.

{Lastly, we reviewed LLM outputs with domain experts for recurring errors, and worked out corresponding fixes. First, Clinic B exhibited substantial missingness in dosage instructions (SIG) and daily quantity, as well as gaps in total prescribed quantity (about 20\% records with key missing entries); to enable downstream calculations, we imputed daily quantity using the typical value observed for the same drug name and imputed total quantity with the median value. Second, while most medications are daily oral tablets/films with explicit instructions, about 5\% prescriptions in Clinic C and E are administered monthly or weekly, because they are  extended-interval injections (e.g., Vivitrol). LLMs frequently failed to translate these schedules into the structured output of daily and total quantities, so we manually set duration based on the documented dosing schedule for these cases. Third, we found unit-related outliers in the LLM outputs: whereas daily quantity is usually in tablets/films, occasional entries listed mass units (e.g., 250 g), which the models misread as a daily count (e.g., 250). We addressed this by adding unit normalization and plausibility checks to prevent misinterpretation and cap extreme values.}

\section{Concluding Remarks}
\label{sec:disc}

In this work, we designed and evaluated a specialized framework to automate the extraction of structured medication data from heterogeneous EHR records, using this information to compute a standardized medication coverage duration (MOUD days). Our results demonstrate that a framework leveraging modern open-source LLMs with constrained JSON generation can serve as a robust and scalable alternative to traditional, brittle ETL pipelines. We found that models capable of running on a single GPU, achieved high accuracy, successfully harmonizing prescription data across five disparate clinical systems without site-specific engineering.

This work provides a practical demonstration of how locally-deployable generative AI can address critical data exchange and integration problems in healthcare. By enabling the rapid, automated calculation of standardized metrics like MOUD days, our framework can significantly accelerate multi-site EHR research. This allows clinicians and policymakers to better understand treatment adherence and retention, particularly in the underserved rural communities highlighted in this study.

\paragraph{Limitations}
Despite these promising results, we acknowledge several limitations. First, while tested across five diverse clinics, the framework's generalizability to a wider range of EHR systems and documentation styles remains to be validated. We anticipate that newer reasoning models will handle heterogeneous EHRs with less prompt customization; evaluating this hypothesis is a priority for future work.
Second, our current implementation focuses solely on MOUD prescriptions; extending it to other medication classes or clinical concepts would require further prompt engineering and evaluation. Finally, like all LLM-based systems,  for high-stakes clinical applications, a human-in-the-loop review process would be essential to validate the extracted data.

\paragraph{Future Work}
Future work will address these limitations. We plan to expand our evaluation to a larger group of clinics and benchmark against reasoning-focused models \cite{guo2025deepseek,yang2025qwen3}, which have shown strong performance on verifiable tasks like math and coding. We will also explore efficient fine-tuning techniques, such as LoRA \cite{lora, qlora}, to create smaller, specialized models. Finally, developing a user-friendly interface for expert review and error adjudication is a key priority to prepare the framework for clinical use.

\section{Acknowledgment}
\label{sec:ack}
Research reported in this publication was supported by the National Institute On Drug Abuse of the National Institutes of Health under Award Number UG1DA049435. The content is solely the responsibility of the authors and does not necessarily represent the official views of the National Institutes of Health.

\medskip
{
\small
\bibliographystyle{plainnat} 
\bibliography{custom} 
}

\definecolor{mycolor}{RGB}{230, 245, 255}
\definecolor{orange}{RGB}{255, 240, 204}
\definecolor{colorgreen}{RGB}{240, 255, 220}

\newpage

\appendix
Our technical appendix is structured as follows:
\begin{enumerate}
    \item Appendix~\ref{app:sec:prescription_prompts}: Prompts and Schemas for Prescription Information Extraction
    \item Appendix~\ref{app:sec:model_specs}: Model Specifications and Parameters
    \item Appendix~\ref{app:sec:inference_config}: Inference Configuration and Settings
\end{enumerate}

\newpage

\section{Prompts and Schemas for Prescription Information Extraction}
\label{app:sec:prescription_prompts}

This section details the system prompt, format instructions, and Pydantic output schemas used for extracting structured information from various prescription formats.

\subsection{General System Prompt}
All extraction tasks are guided by the following high-level instruction, which establishes the persona and objective for the LLM.
\begin{tcolorbox}[colback=mycolor,colframe=blue!50!black,title=System Prompt]
\begin{lstlisting}[basicstyle=\ttfamily, breaklines=true]
You are a medical expert, you are tasked with extracting useful information from a prescription. Before answering you should reason about the problem (using the "reasoning" field in the JSON response). You need to follow the format described below:
\end{lstlisting}
\end{tcolorbox}

\subsection{Shared Extraction Rules and Guidelines}
Across all data formats, a common set of interpretation and post-processing rules are applied, as specified in the "Important Notes" section of each prompt.

\begin{tcolorbox}[colback=mycolor,colframe=blue!50!black,title=Important Notes:]
\begin{lstlisting}[basicstyle=\ttfamily, breaklines=true]
- If information is unavailable, set the field to None.
- Convert fractions with space (e.g., '3 1/2') to decimal values (e.g., '3.5').
- For medication frequency: interpret "X10" as 10 days, but only when X is followed by a reasonable number. Don't apply this rule if P follows X or if the number is unusually large.
- Always include detailed step-by-step calculations in the "reasoning" field, particularly for injections and complex dosing regimens.
- Watch for specialized dosing terms: "inject/injection," "patch," "every 4 weeks," "monthly," "weekly," "once a week," "every 7 days," etc.
- For injection medications, carefully analyze the SIG field to determine proper administration schedule.
- Special medications like Vivitrol are injections administered monthly - always note this in your reasoning.
- When extracting daily quantities from dosage instructions, sum all individual doses (e.g., "one tab in morning, half tab at night" = 1.5).
- For duration calculations, extract explicit day counts or convert frequency information (weekly = 7 days, monthly = 30 days, etc.).
- Convert text numbers to numerals: "one" -> 1, "two" -> 2, etc.
\end{lstlisting}
\end{tcolorbox}

\newpage

\subsubsection{Hometown Prescription Format}

\noindent\textbf{Prompt Format:}
\begin{tcolorbox}[colback=orange!20,colframe=orange!80!black]
\begin{lstlisting}[basicstyle=\ttfamily, breaklines=true]
{
    "reasoning": <reasoning about the answer>,
    "patient_id": <extract from 'RECORDID'>,
    "prescription_date": <extract from 'PRESCRIPTION_DATE'>,
    "drug_name": <extract from 'GENERIC_NAME', without usage info>,
    "drug_name_full": <extract from 'GENERIC_NAME'>,
    "total_quantity": <extract from 'PRESCRIBED_QUANTITY'>,
    "daily_quantity": <calculate from DOSAGE_INSTRUCTION>,
    "Refill": <extract from 'ORIGINAL_REFILLS'>,
    "drug_strength": <extract from 'UNIT_DOSE'>,
    "drug_form": <extract from 'ROUTE_OF_ADMINISTRATION'>,
    "SIG": <extract from 'DOSAGE_INSTRUCTIONS'>,
    "prescriber_id": <extract from 'PRESCRIBER_ID'>
}
\end{lstlisting}
\end{tcolorbox}

\noindent\textbf{Pydantic Schema: \texttt{HometownPrescription}}
\vspace{1ex}

\begingroup
\small
\setlength{\tabcolsep}{6pt}
\renewcommand{\arraystretch}{1.2}

\begin{tabularx}{\linewidth}{@{} >{\raggedright\arraybackslash}p{0.26\linewidth} >{\raggedright\arraybackslash\ttfamily}p{0.22\linewidth} >{\raggedright\arraybackslash}X @{}}
\toprule
\textbf{Field} & \textbf{Type} & \textbf{Description} \\
\midrule
reasoning & Optional[str] & Reasoning about the answer. \\
patient\_id & Optional[str] & Extract from \texttt{RECORDID}. \\
prescription\_date & Optional[str] & Extract from \texttt{PRESCRIPTION\_DATE}. \\
drug\_name & Optional[str] & Extract from \texttt{GENERIC\_NAME}, without usage info. \\
drug\_name\_full & Optional[str] & Extract from \texttt{GENERIC\_NAME}. \\
total\_quantity & Optional[float] & Extract from \texttt{PRESCRIBED\_QUANTITY}. \\
daily\_quantity & Optional[float] & Calculate from \texttt{DOSAGE\_INSTRUCTION} by summing quantities. \\
Refill & Optional[float] & Extract from \texttt{ORIGINAL\_REFILLS}. \\
drug\_strength & Optional[str] & Extract from \texttt{UNIT\_DOSE}. \\
drug\_form & Optional[str] & Extract from \texttt{ROUTE\_OF\_ADMINISTRATION}. \\
SIG & Optional[str] & Directions for use. Extract from \texttt{DOSAGE\_INSTRUCTIONS}. \\
prescriber\_id & Optional[str] & Extract from \texttt{PRESCRIBER\_ID}. \\
\bottomrule
\end{tabularx}
\endgroup

\newpage

\subsubsection{Providence Prescription Format}

\noindent\textbf{Prompt Format:}
\begin{tcolorbox}[colback=orange!20,colframe=orange!80!black]
\begin{lstlisting}[basicstyle=\ttfamily, breaklines=true]
{
    "reasoning": <reasoning about the answer>,
    "patient_id": <extract from 'record_num'>,
    "prescription_date": <extract from 'order_date'>,
    "drug_name": <extract from 'epic_medication_name', without usage info>,
    "drug_name_full": <extract from 'epic_medication_name'>,
    "total_quantity": <extract from 'dose_instructions'>,
    "daily_quantity": <extract from 'dose_instructions'>,
    "Refill": <number of refills>,
    "duration": <extract days from 'dose_instructions' or set NA>,
    "drug_strength": <extract from 'epic_medication_name'>,
    "drug_form": <extract from 'dose_unit'>,
    "SIG": <extract from 'dose_instructions'>,
    "prescriber_id": <extract from 'prescriber_id'>
}
\end{lstlisting}
\end{tcolorbox}

\noindent\textbf{Pydantic Schema: \texttt{ProvidencePrescription}}
\vspace{1ex}

\begingroup
\small
\setlength{\tabcolsep}{6pt}
\renewcommand{\arraystretch}{1.2}

\begin{tabularx}{\linewidth}{@{} >{\raggedright\arraybackslash}p{0.26\linewidth} >{\raggedright\arraybackslash\ttfamily}p{0.22\linewidth} >{\raggedright\arraybackslash}X @{}}
\toprule
\textbf{Field} & \textbf{Type} & \textbf{Description} \\
\midrule
reasoning & Optional[str] & Reasoning about the answer. \\
patient\_id & Optional[str] & Extract from \texttt{record\_num}. \\
prescription\_date & Optional[str] & Extract from \texttt{order\_date}. \\
drug\_name & Optional[str] & Extract from \texttt{epic\_medication\_name}, without usage info. \\
drug\_name\_full & Optional[str] & Extract from \texttt{epic\_medication\_name}. \\
total\_quantity & Optional[float] & Extract from \texttt{dose\_instructions}. \\
daily\_quantity & Optional[float] & Extract from \texttt{dose\_instructions}. \\
Refill & Optional[float] & Number of refills prescribed. \\
duration & Optional[float] & Extract days from \texttt{dose\_instructions} if present. \\
drug\_strength & Optional[str] & Extract from \texttt{epic\_medication\_name}. \\
drug\_form & Optional[str] & Extract from \texttt{dose\_unit}. \\
SIG & Optional[str] & Directions for use. Extract from \texttt{dose\_instructions}. \\
prescriber\_id & Optional[str] & Extract from \texttt{prescriber\_id}. \\
\bottomrule
\end{tabularx}
\endgroup

\newpage

\subsubsection{Seaport Prescription Format}

\noindent\textbf{Prompt Format:}
\begin{tcolorbox}[colback=orange!20,colframe=orange!80!black]
\begin{lstlisting}[basicstyle=\ttfamily, breaklines=true]
{
    "reasoning": <reasoning about the answer>,
    "patient_id": <extract from 'Record_ID'>,
    "prescription_date": <extract from 'Prescription_Date'>,
    "drug_name": <extract from 'RX_Name', without usage info>,
    "drug_name_full": <extract from 'RX_Name'>,
    "total_quantity": <extract from 'Quantity'>,
    "daily_quantity": <extract from 'Unit_Dose' or 'SIG'>,
    "Refill": <number of refills>,
    "duration": <extract days from 'SIG' or set NA>,
    "drug_strength": <extract from 'RX_Name'>,
    "drug_form": <extract from 'Display_Dosage_Unit'>,
    "SIG": <extract from 'SIG'>,
    "prescriber_id": <extract from 'PRESCRIBER_ID'>
}
\end{lstlisting}
\end{tcolorbox}

\noindent\textbf{Pydantic Schema: \texttt{SeaportPrescription}}
\vspace{1ex}

\begingroup
\small
\setlength{\tabcolsep}{6pt}

\renewcommand{\arraystretch}{1.2}
\begin{tabularx}{\linewidth}{@{} >{\raggedright\arraybackslash}p{0.26\linewidth} >{\raggedright\arraybackslash\ttfamily}p{0.22\linewidth} >{\raggedright\arraybackslash}X @{}}
\toprule
\textbf{Field} & \textbf{Type} & \textbf{Description} \\
\midrule
reasoning & Optional[str] & Reasoning about the answer. \\
patient\_id & Optional[str] & Extract from \texttt{Record\_ID}. \\
prescription\_date & Optional[str] & Extract from \texttt{Prescription\_Date}. \\
drug\_name & Optional[str] & Extract from \texttt{RX\_Name}, without usage info. \\
drug\_name\_full & Optional[str] & Extract from \texttt{RX\_Name}. \\
total\_quantity & Optional[float] & Extract from \texttt{Quantity}. \\
daily\_quantity & Optional[float] & Extract from \texttt{Unit\_Dose} or \texttt{SIG}. \\
Refill & Optional[float] & Number of refills prescribed. \\
duration & Optional[float] & Extract days from \texttt{SIG} if present. \\
drug\_strength & Optional[str] & Extract from \texttt{RX\_Name}. \\
drug\_form & Optional[str] & Extract from \texttt{Display\_Dosage\_Unit}. \\
SIG & Optional[str] & Directions for use. Extract from \texttt{SIG}. \\
prescriber\_id & Optional[str] & Extract from \texttt{PRESCRIBER\_ID}. \\
\bottomrule
\end{tabularx}
\endgroup

\newpage

\subsubsection{St. Mary's Prescription Format}

\noindent\textbf{Prompt Format:}
\begin{tcolorbox}[colback=orange!20,colframe=orange!80!black]
\begin{lstlisting}[basicstyle=\ttfamily, breaklines=true]
{
    "reasoning": <reasoning about the answer>,
    "patient_id": <extract from 'RecordNumber'>,
    "prescription_date": <extract from 'PrescriptionDate'>,
    "drug_name": <extract from 'Description', without usage info>,
    "drug_name_full": <extract from 'Description'>,
    "total_quantity": <extract from 'DoseQuantity'>,
    "daily_quantity": <extract from 'UnitDosage' or 'DosageInstructions'>,
    "Refill": <number of refills>,
    "duration": <extract days from 'DosageInstructions' or set NA>,
    "drug_strength": <extract from 'Description'>,
    "drug_form": <extract from 'Description'>,
    "SIG": <extract from 'DosageInstructions'>,
    "prescriber_id": <extract from 'PrescriberID'>
}
\end{lstlisting}
\end{tcolorbox}

\noindent\textbf{Pydantic Schema: \texttt{StMarysPrescription}}
\vspace{1ex}

\begingroup
\small
\setlength{\tabcolsep}{6pt}
\renewcommand{\arraystretch}{1.2}

\begin{tabularx}{\linewidth}{@{} >{\raggedright\arraybackslash}p{0.26\linewidth} >{\raggedright\arraybackslash\ttfamily}p{0.22\linewidth} >{\raggedright\arraybackslash}X @{}}
\toprule
\textbf{Field} & \textbf{Type} & \textbf{Description} \\
\midrule
reasoning & Optional[str] & Reasoning about the answer. \\
patient\_id & Optional[str] & Extract from \texttt{RecordNumber}. \\
prescription\_date & Optional[str] & Extract from \texttt{PrescriptionDate}. \\
drug\_name & Optional[str] & Extract from \texttt{Description}, without usage info. \\
drug\_name\_full & Optional[str] & Extract from \texttt{Description}. \\
total\_quantity & Optional[float] & Extract from \texttt{DoseQuantity}. \\
daily\_quantity & Optional[float] & Extract from \texttt{UnitDosage} or \texttt{DosageInstructions}. \\
Refill & Optional[float] & Number of refills prescribed. \\
duration & Optional[float] & Extract days from \texttt{DosageInstructions} if present. \\
drug\_strength & Optional[str] & Extract from \texttt{Description}. \\
drug\_form & Optional[str] & Extract from \texttt{Description}. \\
SIG & Optional[str] & Directions for use. Extract from \texttt{DosageInstructions}. \\
prescriber\_id & Optional[str] & Extract from \texttt{PrescriberID}. \\
\bottomrule
\end{tabularx}
\endgroup

\newpage

\subsubsection{Syringa Prescription Format}

\noindent\textbf{Prompt Format:}
\begin{tcolorbox}[colback=orange!20,colframe=orange!80!black]
\begin{lstlisting}[basicstyle=\ttfamily, breaklines=true]
{
    "reasoning": <reasoning about the answer>,
    "patient_id": <extract from 'Record Number'>,
    "prescription_date": <extract from 'Order Dt/Tm'>,
    "drug_name": <extract from 'Order Mnemonic', without usage info>,
    "drug_name_full": <extract from 'Order Mnemonic'>,
    "total_quantity": <extract from 'Dispense Qty'>,
    "daily_quantity": <extract from 'Volume Dose'>,
    "Refill": <number of refills>,
    "Frequency": <extract from 'Frequency': daily=1, BID=2, TID=3>,
    "drug_form": <extract from 'Volume Dose Unit'>,
    "SIG": <extract from 'Frequency'>,
    "prescriber_id": <extract from 'Order Last Updt Provider Id'>
}
\end{lstlisting}
\end{tcolorbox}

\noindent\textbf{Pydantic Schema: \texttt{SyringaPrescription}}
\vspace{1ex}

\begingroup
\small
\setlength{\tabcolsep}{6pt}
\renewcommand{\arraystretch}{1.2}

\begin{tabularx}{\linewidth}{@{} >{\raggedright\arraybackslash}p{0.26\linewidth} >{\raggedright\arraybackslash\ttfamily}p{0.22\linewidth} >{\raggedright\arraybackslash}X @{}}
\toprule
\textbf{Field} & \textbf{Type} & \textbf{Description} \\
\midrule
reasoning & Optional[str] & Reasoning about the answer. \\
patient\_id & Optional[str] & Extract from \texttt{Record Number}. \\
prescription\_date & Optional[str] & Extract from \texttt{Order Dt/Tm}. \\
drug\_name & Optional[str] & Extract from \texttt{Order Mnemonic}, without usage info. \\
drug\_name\_full & Optional[str] & Extract from \texttt{Order Mnemonic}. \\
total\_quantity & Optional[float] & Extract from \texttt{Dispense Qty}. \\
daily\_quantity & Optional[float] & Extract from \texttt{Volume Dose}. \\
Refill & Optional[float] & Number of refills prescribed. \\
Frequency & Optional[float] & Extract from \texttt{Frequency} (daily{=}1, BID{=}2, etc.). \\
drug\_form & Optional[str] & Extract from \texttt{Volume Dose Unit}. \\
SIG & Optional[str] & Directions for use. Extract from \texttt{Frequency}. \\
prescriber\_id & Optional[str] & Extract from \texttt{Order Last Updt Provider Id}. \\
\bottomrule
\end{tabularx}
\endgroup

\newpage

\subsubsection{Winterport Prescription Format}

\noindent\textbf{Prompt Format:}
\begin{tcolorbox}[colback=orange!20,colframe=orange!80!black]
\begin{lstlisting}[basicstyle=\ttfamily, breaklines=true]
{
    "reasoning": <reasoning about the answer>,
    "patient_id": <extract from 'RecordID'>,
    "prescription_date": <extract from 'PrescribedDate'>,
    "drug_name": <extract from 'DrugDescription', without usage info>,
    "drug_name_full": <extract from 'DrugDescription'>,
    "total_quantity": <extract from 'PrescribedQuantity'>,
    "daily_quantity": <extract from 'DoseQuantity' or 'INSTRUCTIONS'>,
    "Refill": <number of refills>,
    "duration": <extract days from 'INSTRUCTIONS' or set NA>,
    "drug_strength": <extract from 'DrugDescription'>,
    "drug_form": <extract from 'ROUTE'>,
    "SIG": <extract from 'INSTRUCTIONS'>,  
    "prescriber_id": <extract from 'Prescriber'>
}
\end{lstlisting}
\end{tcolorbox}


\noindent\textbf{Pydantic Schema: \texttt{WinterportPrescription}}
\vspace{1ex}

\begingroup
\small
\setlength{\tabcolsep}{6pt}
\renewcommand{\arraystretch}{1.2}

\begin{tabularx}{\linewidth}{@{} >{\raggedright\arraybackslash}p{0.26\linewidth} >{\raggedright\arraybackslash\ttfamily}p{0.22\linewidth} >{\raggedright\arraybackslash}X @{}}
\toprule
\textbf{Field} & \textbf{Type} & \textbf{Description} \\
\midrule
reasoning & Optional[str] & Reasoning about the answer. \\
patient\_id & Optional[str] & Extract from \texttt{RecordID}. \\
prescription\_date & Optional[str] & Extract from \texttt{PrescribedDate}. \\
drug\_name & Optional[str] & Extract from \texttt{DrugDescription}, without usage info. \\
drug\_name\_full & Optional[str] & Extract from \texttt{DrugDescription}. \\
total\_quantity & Optional[float] & Extract from \texttt{PrescribedQuantity}. \\
daily\_quantity & Optional[float] & Extract from \texttt{DoseQuantity} or \texttt{INSTRUCTIONS} by summing. \\
Refill & Optional[float] & Number of refills prescribed. \\
duration & Optional[float] & Extract days from \texttt{INSTRUCTIONS} if present. \\
drug\_strength & Optional[str] & Extract from \texttt{DrugDescription}. \\
drug\_form & Optional[str] & Extract from \texttt{ROUTE}. \\
SIG & Optional[str] & Directions for use. Extract from \texttt{INSTRUCTIONS}. \\
prescriber\_id & Optional[str] & Extract from \texttt{Prescriber}. \\
\bottomrule
\end{tabularx}
\endgroup

\newpage

\section{Model Specifications and Parameters}
\label{app:sec:model_specs}

\subsection{Qwen Model Family}
\label{app:sub:qwen_family}

\textbf{Qwen2.5 (32B, GPTQ)}
\begin{itemize}
    \item \textbf{Architecture}: Transformer-based decoder-only architecture with RoPE positional embeddings, SwiGLU, RMSNorm, attention with QKV bias
    \item \textbf{Parameters}: 32.5 billion parameters
    \item \textbf{Quantization}: GPTQ 8-bit quantization
    \item \textbf{Context Length}: 131,072 tokens (native support), generation typically up to 8,000 tokens
    \item \textbf{Pre-training}: General domain corpus (no public confirmation of medical data)
\end{itemize}

\textbf{Qwen3 (32B)}
\begin{itemize}
    \item \textbf{Architecture}: Enhanced transformer decoder with GQA (Grouped Query Attention), SwiGLU, RMSNorm, RoPE
    \item \textbf{Parameters}: 32.8 billion parameters
    \item \textbf{Quantization}: GGUF 4 bit.
    \item \textbf{Context Length}: 128,000 tokens (with YaRN scaling)
    
\end{itemize}

\textbf{Qwen3 (8B)}
\begin{itemize}
    \item \textbf{Parameters}: 8.2 billion parameters (6.95B non-embedding)
    \item \textbf{Quantization}: No quantization, BF16 inference.
    \item \textbf{Context Length}: 32,768 tokens (extendable to 131,072 with YaRN)
\end{itemize}

\textbf{Qwen3 (4B)}
\begin{itemize}
    \item \textbf{Parameters}: 4.8 billion parameters
    \item \textbf{Quantization}: No quantization, BF16 inference.
    \item \textbf{Context Length}: 32,768 tokens (YaRN extendable)
\end{itemize}

\subsection{Gemma Model Family}
\label{app:sub:gemma_family}

\textbf{Gemma 3 (27B)}
\begin{itemize}
    \item \textbf{Architecture}: Transformer-based decoder-only with grouped-query attention (GQA) and SigLIP vision encoder; multilingual (140+ languages), multimodal (text + image)
    \item \textbf{Parameters}: 27 billion parameters
    \item \textbf{Quantization}: No quantization, BF16 inference. 
    \item \textbf{Context Length}: 128,000 tokens (long context support)

\end{itemize}

\textbf{Gemma 3 (4B)}
\begin{itemize}
    \item \textbf{Parameters}: 4 billion parameters
    \item \textbf{Quantization}: No quantization, BF16 inference.
    \item \textbf{Context Length}: 128,000 tokens
    \item \textbf{Multimodal Capability}: Supports both text and image inputs
    \item \textbf{Pre-training}: Same distilled training approach as 27B
\end{itemize}

\subsection{MedGemma Model Family}
\label{app:sub:medgemma_family}

\textbf{MedGemma (27B)}
\begin{itemize}
    \item \textbf{Parameters}: 27 billion parameters
    \item \textbf{Quantization}: No quantization, BF16 inference.
    \item \textbf{Context Length}: 128,000 tokens
    \item \textbf{Medical Pre‑training}: Includes medical text (EHRs, question-answer pairs), FHIR-format clinical records, and medical images (e.g. chest X-ray, pathology, ophthalmology, dermatology)
    \item \textbf{Intended Use}: Research and development in healthcare AI; not clinical-grade; part of Health AI Developer Foundations
\end{itemize}

\textbf{MedGemma (4B)}
\begin{itemize}
    \item \textbf{Parameters}: 4 billion parameters
    \item \textbf{Quantization}: No quantization, BF16 inference.
    \item \textbf{Context Length}: 128,000 tokens
    \item \textbf{Medical Pre‑training}: Same data modalities as MedGemma 27B (medical text + images)
    \item \textbf{Intended Use}: Lightweight deployment and health‑AI prototyping; not clinical use
\end{itemize}

\newpage
\section{Inference Configuration and Settings}
\label{app:sec:inference_config}

\subsection{Hardware Configuration}
\label{app:sub:hardware_config}
\begin{itemize}
    \item \textbf{Primary GPUs}: 4× NVIDIA A6000 (48GB VRAM each, 192GB total)
    \item \textbf{CPU}: Intel Xeon Gold 6448Y 64-Core Processor
    \item \textbf{System Memory}: 512GB RAM
\end{itemize}

\subsection{vLLM Configuration}
\label{app:sub:vllm_config}
\begin{verbatim}
vllm_config:
  model_name: "model-specific"
  pipeline_parallel_size: 1
  tensor_parallel_size: 2
  gpu_memory_utilization: 0.95
  dtype: "auto"
  quantization: "gptq"  # for applicable models
  trust_remote_code: true
  seed: 42
\end{verbatim}

\subsubsection{Inference Parameters}
\label{app:sub:inference_params}
\begin{itemize}
    \item \textbf{Temperature}: 0 (for consistency)
    \item \textbf{Max New Tokens}: 4092
    \item \textbf{Batch Size}: 
        \begin{itemize}
            \item 32B models: 200 
            \item 8B models: 500
            \item 4B models: 500
        \end{itemize}
    \item \textbf{Output Format}: JSON with Pydantic validation
\end{itemize}

\subsubsection{Optimization Techniques}
\label{app:sub:optimization}
\begin{itemize}
    \item \textbf{Quantization Methods}:
        \begin{itemize}
            \item GPTQ: 8-bit quantization
            \item GGUF: 4-bit quantization
        \end{itemize}
    \item \textbf{Memory Management}:
        \begin{itemize}
            \item Gradient checkpointing (when applicable)
            \item KV-cache optimization
        \end{itemize}
    \item \textbf{Throughput Optimization}:
        \begin{itemize}
            \item Continuous batching in vLLM
            \item PagedAttention for memory efficiency
            \item Tensor parallelism for large models
        \end{itemize}
\end{itemize}

\end{document}